\documentclass{article}
\usepackage{graphicx} 
\usepackage{hyperref}

\begin{document}

\title{AI Safety: Necessary, but insufficient and possibly problematic\footnote{Expanded version of article in AI \& Society journal \url{https://link.springer.com/article/10.1007/s00146-024-01899-y}}}
\author{Deepak P. \\ Queen's University Belfast, UK \\ deepaksp@acm.org}
\date{}

\maketitle

\begin{abstract}
This article critically examines the recent hype around AI safety. We first start with noting the nature of the AI safety hype as being dominated by governments and corporations, and contrast it with other avenues within AI research on advancing social good. We consider what 'AI safety' actually means, and outline the dominant concepts that the digital footprint of AI safety aligns with. We posit that AI safety has a nuanced and uneasy relationship with transparency and other allied notions associated with societal good, indicating that it is an insufficient notion if the goal is that of societal good in a broad sense. We note that the AI safety debate has already influenced some regulatory efforts in AI, perhaps in not so desirable directions. We also share our concerns on how AI safety may normalize AI that advances structural harm through providing exploitative and harmful AI with a veneer of safety. 
\end{abstract}


\subsubsection*{Global Clamor for AI Safety} 

There has been a newfound global enthusiasm around AI safety. This has notably been at the behest of global governments and corporations, and with only limited involvement by the academic and scholarly community on AI across global universities. In what would seem to an onlooker like a global coordinated action over Nov-Dec 2023, the series started with the AI safety summit\footnote{\url{https://www.aisafetysummit.gov.uk/}} organized by the UK which attracted participation from across 28 nations and big tech bosses such as Musk and Altman. Around the same time, an AI safety executive order~\cite{execorder} from the White House that mandated that AI developers should share safety results with government was issued. The above two nations quickly followed it up by announcing their own AI safety institutes, the UK institute within the Department of Science Innovation and Technology\footnote{\url{https://www.gov.uk/government/organisations/ai-safety-institute}} and the US institute within the NIST,\footnote{\url{https://www.nist.gov/artificial-intelligence/artificial-intelligence-safety-institute}}. France and South Korea made sure they were not left behind by signing up to host separate global AI safety summits\footnote{\url{https://www.reuters.com/technology/south-korea-france-host-next-two-ai-safety-summits-2023-11-01/}}, both slated to take place in 2024. The Global Partnership in AI (GPAI) hosted a ministerial summit of 29 nations in the second week of December 2023 at New Delhi, with the declaration~\cite{gpaiministerial} emerging out of it also using safe as the first adjective in their ideal of AI. The GPAI declaration contrasted with the AI safety declaration emerging out of the UK AI safety summit in being more pro-innovation, with the language making it sound like being more ‘balanced’ in the trade-off between innovation and safety\footnote{\url{https://indianexpress.com/article/explained/explained-sci-tech/delhi-declaration-gpai-regulation-ai-explained-9067865/}}, implicitly suggesting that some compromises on safety may be necessary. In this context, it is also interesting to note that the EU AI act~\cite{euaiact}, borne out of discussions that started prior to the AI safety hoopla, focuses on differentiating AI on risk levels, striking a distinctly different tone than safety in various parts. 

\subsubsection*{The AI Scholarly Community and Social Good}

The aforementioned enthusiasm around AI safety may make citizens think that governments and corporations have leapfrogged the AI scholarly community in imagining AI as a force for social good. This is, however, hardly the case. The scholarly community in AI has, over the past decade, seen much enthusiasm and brisk research around AI for social good and allied directions. The terms which have been adjectivized for such purposes include ethics, equity, responsibility, trust, security, explainability and accountability, among others. The flagship professional community in computing, ACM, instituted no less than three conferences on the theme in the past decade. While the ACM conference on Fairness, Accountability and Transparency (FAccT\footnote{\url{https://facctconference.org/}}, earlier FAT*) and the AAAI/ACM Conference on AI Ethics and Society (AIES\footnote{\url{https://www.aies-conference.com/}}) started as annual events in 2018, the ACM Conference on Equity and Access in Algorithms, Mechanisms, and Optimization (EEAMO\footnote{\url{https://eaamo.org/}}) and ACM Conference on Information Technology for Social Good (GoodIT\footnote{\url{https://dl.acm.org/conference/goodit}}) both held their first editions in 2021. Among journals, the AI \& Society journal\footnote{\url{https://link.springer.com/journal/146}} has been operating for more than three decades, and new publications such as AI \& ethics journal\footnote{\url{https://link.springer.com/journal/43681}}  (since 2021) and Critical AI journal\footnote{\url{https://criticalai.org/}} (since 2023) entered the fray in the past years. While this list is far from comprehensive, it serves to illustrate that the AI safety buzz is divergent from the themes in AI scholarship in envisioning social good from AI. 

\subsubsection*{Unpacking 'AI Safety'}

Against the backdrop outlined, let us try to unpack at AI safety. AI safety is potentially an attractive term that comes across also as a no-brainer; we would all, undoubtedly, like AI to be safe! Safety has a literal meaning within its usual usage in non-technical contexts, one that aligns with a kind of operation where nothing {\it unexpected} or {\it undesirable} happens {\it visibly}. We posit that the current global enthusiasm on AI safety is intricately correlated with such notions. Let us consider these notions in turn.

The narrowest interpretation of the notion of safety may be that of avoiding unexpected kinds of operation. Complex programs going awry is nothing new in the practice of software development, and the traditional mechanism of doing away with unexpected operation is to institute a comprehensive quality control ({\it aka} software testing) process. Once all ‘bugs’ – defined as errors, flaws or faults in the design, development or operation of the software – are identified through testing and weeded out, the software may, to a high confidence, be expected to work as intended. Interpreting AI safety as asserting that software should work as expected could be regarded as somewhat naïve, since it is simply a restatement of the well understood necessity of quality control in software design, development and maintenance. It is not clear whether this interpretation of safety as quality control is envisaged within the remit of AI safety – yet, it is noteworthy that the UK’s AI safety institute policy paper laments that {\it ‘ … there is no common standard in quality or consistency’}. 

A somewhat stronger interpretation of AI Safety along the same direction is to ensure that the AI be not nudged to work in unexpected ways by malicious actors such as hackers. In these days when most AI operate within networks of computers, robustness to malicious hackers is an increasing concern. This has spawned a bustling field of activity, called cybersecurity. This stronger interpretation of AI safety insists that software is not just well-tested for normal operation, but are also able to prevent, resist or otherwise be robust to unexpected actions from malicious actors. One popular way of doing this is through a paradigm called ‘red teaming’, whereby a dedicated in-house (red) team are asked to interact with a software adversarially, to uncover potential vulnerabilities. These vulnerabilities can then be corrected prior to rollout. In the latest wave of generative AI, red teaming has started becoming popular (e.g., Ganguli et al, 2022~\cite{ganguli2022red}) as a way of embedding robustness to malicious actors, within AI. The interpretation of AI safety from US may be interpreted as including this notion of robustness to malicious actors; for instance, two of the working groups under the remit of the US AI safety institute\footnote{\url{https://www.nist.gov/artificial-intelligence/artificial-intelligence-safety-institute}} are titled {\it ‘red-teaming’} and {\it ‘safety \& security’}. 

We now come to the case of undesirable operation, which we examine in tandem with the notion of visibility. AI usage within very common contemporary scenarios, such as consumer services (e.g., social networks, chatbots, information personalization, recommenders), public sector contexts (e.g., welfare application processing, policing, healthcare delivery) and enterprise contexts (e.g., resource planning, hiring, processing of credit applications, enterprise knowledge bases) often involve making substantive decisions affecting humans and societies. It would be important to consider what would be undesirable kinds of operation in such cases. To keep things grounded, we consider a concrete example to anchor the discussion. A person whose job application has been rejected by AI due to unacceptable reasons (reasons implicitly – but not necessarily apparently – aligned with gender or race, or other similar) is unquestionably harmed, since the public consensus within the contemporary world is that individuals be not victimized based on their protected characteristics\footnote{\url{https://www.equalityhumanrights.com/equality/equality-act-2010/protected-characteristics}}. Yet, what kind of harm is it? ? It may not be a quality control deficit, since the current AI may not have been – nor expected to be – explicitly instructed to profile all possible reasons for correlations with protected attributes such as gender or race. It is not evidently a security issue either. It is a harm of misalignment (Ji et al., 2023~\cite{ji2023ai}) where the AI has not sufficiently internalized the contemporary public consensus on ethics and values. Misalignment and consequent structural harm are clearly undesirable, but how do we catch it? We may need to rely on the user to point out that she has been disadvantaged. This may not be possible if the AI does not offer a rationale for its decision in the first place. The complainant usually needs to bear the burden of proof, but the proof may not be visible from the AI itself, making it impossible to complain in the first place. If the AI offers an explanation that the decision was made based on factors that are correlated with protected characteristics, it could be caught. Yet, since ‘AI safety’ does not apparently mandate transparency, structural harm due to AI could still be passed off under the remit of ‘safe AI’ as long as it is not patently visible. If AI safety is primed towards addressing visible harm, this begs the question as to where AI safety stands with respect to mandating visibility in AI. 

\subsubsection*{AI Safety vis-a-vis Transparency}

It is noteworthy that the US executive order issued at the beginning of the emergence of the AI safety movement stresses on a kind of transparency\footnote{\url{https://www.bbc.co.uk/news/technology-67261284}}, that of sharing {\it safety results} with the {\it government} (cf. public). This is an inadequate notion of transparency if the intent is to tackle misalignment. There could be at least three kinds of transparency. First, in what one may call as technological transparency, we may want to mandate that the source code of the AI, as well as the data it is trained on, is made transparent. Second, we may desire that the objectives that are encoded within the AI – e.g., efficiency, profit maximization, reduction of waste – be made transparent. Third, we may insist that every decision made by the AI be supported by justifications, providing transparency at the decision level. Arguably, the first and third are attempted to some extent within EU’s GDPR (Ref: Recital 71\footnote{\url{https://www.privacy-regulation.eu/en/r71.htm}}), but the AI safety debates have hardly gone into any of the above three kinds. There is hardly any mention of enhancing visibility within the AI safety debates. This potentially paves the path for AI developers to adopt the opaqueness route to concur with AI safety than take the harder and potentially economically painful route of bearing the burden of full transparency. It is also notable that taking up full transparency voluntarily comes with the additional risk of opening up the AI to public scrutiny leading to potential embarrassment. 

\subsubsection*{AI Safety vis-a-vis Societal Good}

The curious and critical reader may point out that there exist mentions of societal impacts and other values in parts of the literature emerging from the AI safety debates. For example, the motivation part of the UK AI safety institute policy paper~\cite{aisafetyinstuk} laments that AI could {\it ‘concentrate unaccountable power into the hands of a few’} and cause {\it ‘harms to people’}. The 2023 GPAI ministerial declaration~\cite{gpaiministerial} goes a lot further by proclaiming in the opening paragraph about being {\it ‘rooted in democratic values and human rights, safeguarding dignity and well-being, ensuring personal data protection, protection of applicable intellectual property rights, privacy, and security, fostering innovation, and promoting, trustworthy, responsible, sustainable and human-centred use of AI’}. Yet, when it comes to the operational aspects of the proposals from any of the AI safety initiatives, there is hardly any visible effort towards charting a pathway to address such lofty goals. 

\subsubsection*{AI Safety and Influence on Regulation: The EU AI Act}

If notions of AI safety, as emerging and crystallizing through the global debate, are grossly insufficient to address structural and invisible harm, where does that lead us to. It is interesting to note that there is some latent influence that the enthusiasm of ‘AI safety’ has exerted on regulatory efforts on AI. As a case in point, the trajectory of the EU AI act is worth noting. The initial draft from April 2021\footnote{\url{https://eur-lex.europa.eu/legal-content/EN/TXT/HTML/?uri=CELEX:52021PC0206}} pre-dated the AI safety hoopla, but had many mentions of the word ‘safety’; however, most of those related to safety of natural persons, with their mentions appearing in the company of words such as health and fundamental rights. The other kind of mention was about the {\it ‘safety components’} of products, which are defined in the same document as {\it '‘safety component of a product or system’ means a component of a product or of a system which fulfills a safety function for that product or system or the failure or malfunctioning of which endangers the health and safety of persons or property’} – thus, this is also in reference to the safety of natural persons as opposed to the kind of safety that AI safety refers to. However, by the time the EU AI regulation was debated upon in December 2023, after the emergence of the AI safety bandwagon, the wording had changed considerably. To quote from a European Parliament press release\footnote{\url{https://www.europarl.europa.eu/news/en/press-room/20231206IPR15699/artificial-intelligence-act-deal-on-comprehensive-rules-for-trustworthy-ai}}, {\it ‘MEPs reached a political deal with the Council on a bill to ensure AI in Europe is safe, respects fundamental rights and democracy, while businesses can thrive and expand’}. The mention of safety here is clearly meant to refer to the AI, than to the safety of natural persons, and arguably the latter has taken a backseat. In a way, the emergence of AI safety as a subject of global imagination has already started influencing regulations of AI towards deprioritizing addressal of structural  and other latent harm towards natural persons from opaque AI. 

\subsubsection*{Risks and Harm under Safe AI}

If indeed the shallow notion of AI safety that allows for continuation of structural harm takes on the front seat in debates about the future of AI, that may lead to many undesirable consequences. AI that is intentionally harmful – such as racist predictive policing algorithms\footnote{\url{https://www.technologyreview.com/2020/07/17/1005396/predictive-policing-algorithms-racist-dismantled-machine-learning-bias-criminal-justice/}}, discriminatory insurance premium determination AI\footnote{\url{https://www.insurancethoughtleadership.com/ai-machine-learning/ai-and-discrimination-insurance}}, poor and unsustainable wages determined by gig AI optimized for platform profits\footnote{\url{https://www.theguardian.com/global-development/2023/may/11/half-of-uk-gig-economy-workers-earn-below-minimum-wage-study-reveals}} – could be fomented under the label of ‘safe AI’. The enthusiasm towards AI safety among the bosses of big tech could be interpreted as an indication that the emerging notion of AI safety is probably not antithetical to their extant domination of the AI scene\footnote{\url{https://www.technologyreview.com/2023/12/05/1084393/make-no-mistake-ai-is-owned-by-big-tech/}}, and could instead help them through providing them with a veneer to carry on their profiteering goals unhindered. To state more provocatively, the contours of the emerging AI safety debate is too risky for humanity in that it may normalize the most potent negative consequence from AI, that of structural harm, and secondly, opaqueness of AI. 

\bibliographystyle{abbrv}
\bibliography{refs}

\end{document}